# Performance of different machine learning methods on activity recognition and pose estimation datasets

Dr. Love Trivedi[†], Raviit Vij[*]
†* Department of Physics and CS, JPIS, Jaipur, Rajasthan, India

**Abstract**

With advancements in computer vision taking place day by day, recently a lot of light is being shed on activity recognition. With the range for real-world applications utilising this field of study increasing across a multitude of industries such as security and healthcare, it becomes crucial for businesses to distinguish which machine learning methods perform better than others in the area. This paper strives to aid in this predicament – building upon previous related work, it employs both classical and ensemble approaches on rich pose estimation (OpenPose) and HAR datasets. Making use of appropriate metrics to evaluate the performance for each model, the results show that overall, random forest yields the highest accuracy in classifying ADLs. Relatively all the models have excellent performance across both datasets, except for logistic regression and AdaBoost which perform poorly in the HAR one. With the limitations of this paper also discussed in the end, the scope for further research is vast, which can use this paper as a base in aims of producing better results.

## 1 Introduction

In the era of AI, there has been a stark increase in the popularity of the interdisciplinary scientific field known as computer vision. It seeks to automate tasks that the human visual system can do by deriving viable information and high-level understanding from digital images and videos. In 2021 alone, the field's market shares were appraised at approximately USD 11.22 billion, with expectations of further expansion at a compound annual growth rate of 7.0% between 2022 and 2023.

This research paper attempts to delve deeper into another area where computer vision is being eminently operated - human pose tracking models. Human pose estimation and tracking is a computer vision task that detects and associates key points within people. Initially, each individual was only distinguished as a bounding box. However, in recent years, computers have started garnering a better understanding while identifying different human body poses, as high-performing algorithms in real-time transform the landscape of computer vision. In this paper, we consider 2D human pose estimation, which works by locating semantic key points from digital images and videos and utilizes the orientation to predict and track a person's or object's location. Accordingly, this allows programs to estimate the spatial positions of a body in its respective digital formats. Some popular 2D human pose estimation methods include OpenPose, HRNet and AlphaPose.

Another area that this paper revolves around is Human Activity Recognition (HAR). This research field aims to identify the actions carried out by an individual by gathering and computing information about the individual's state and the surrounding environment. Though on-body sensors are applied here, their extreme discomfort has led to smartphones emerging as a pioneer in the field. By incorporating inertial sensors such as accelerometers and gyroscopes, information regarding body motion can be retrieved and applied to recognize different human activities. Exploration and comparison for this second dataset are made through methods such as Convolutional Neural Networks (CNN), Logistic Regression, Support Vector Machines, Random Forest and K-nearest neighbour models. In order to properly verify the performance of the models, dynamic and static movements (standing, sitting, walking, running, walking upstairs and walking downstairs) have been considered for classification.

In light of this, it is evident that both human pose estimation and HAR are actively paving the road for many real-life applications, many of them in sports and fitness, surveillance, healthcare and even robotics. Therefore, this paper strives to compare the performance of different machine learning methods on different activity recognition datasets: one is derived from OpenPose and the other from the inbuilt accelerometer and gyroscope sensors of the smartphones used. With this, the research could come to a conjecture on how to apply the results gained from this paper to improve the accuracy for classifying either human poses or recognizing activities for a number of industries.

The rest of the paper is organized as follows. The next section focuses on the analysis in the context of related work. Section 3 outlines the methodology for model implementation. Section 4 mentions the results, with Section 5 discussing and concluding.

## 2 Related Work

In this section, some of the most relevant work pertaining to the methods used to found this research paper is presented. It has been divided into two sub-sections, *pose estimation* and *human action recognition*; while they are closely related, they are generally handled as distinct tasks with frameworks for 2D pose estimation being developed from still images and human action recognition from video sequences.

The literature for activity recognition can be traced back to the late 90s, wherein [1] the distinction between 2D (with and without explicit shape models) and 3D approaches for whole-body or hand motion was made. In [2], the survey focuses on pose-based action recognition methods, with research yielding methods like automatic initialization of human motion, pose estimation, recognition and tracking. These methods have since then, been categorized into two main categories which are '*bottom-up*' and '*top-down*' as discussed in [3]. And yet, [4] presented a tree-structured taxonomy, where methods have been categorized into the 'single' layer approaches and 'hierarchical approaches', with each having several layers of categorization. Although, recognition from static images remains a challenging task, most studies found have been associated with facial recognition or pose estimation techniques. [5] has summarized all the methods for activity recognition from still images and categorized them accordingly.

### 2.1 Pose Estimation

Since one half of this research paper focuses on 2D pose estimation, literature for 3D approaches is not taken into account. The problem of pose estimation has been thoroughly examined over the last decade, from pictorial structures [6, 7] to more recent CNN approaches [8, 9, 10]. It can be seen that there are two primary structures for pose detection – detection based and regression-based methods. Detection based methods target pose estimation as a heat map prediction problem where each pixel in a heat map represents the detection score of a corresponding joint. Exploring the concepts of stacked architecture and multi-scale processing, Newell et al. [11] proposed the Stacked Hourglass Network, which has greatly solved the problem in pose estimation. More methods, variations of the aforementioned, have also been proposed such as Chu et al. [12] suggesting an attention model based on conditional random field (CRF) and Yang et al. [13] replacing the residual unit by a Pyramid Residual Module (PRM).

However, the detection approaches do not provide joint coordinates directly. To recover the pose in (x, y) coordinates, the *argmax* function is usually applied as a post-processing step. On the other hand, regression-based approaches use a nonlinear function that maps the input directly to the desired output – the joint coordinates. Following this example, Toshev and Szegedy et al. [14] proposed their solution based on cascade regression for body part detection with Carreira et al. [15] putting forward the Iterative Error Feedback. The limitation of such methods is that the regression function is frequently sub-optimal. In order to tackle this weakness, the Soft-argmax function [16] has been proposed to convert heat maps directly to joint coordinates and consequently allow detection methods to be transformed into regression methods. The main advantage here for regression methods over detection ones is that they often are fully differentiable. This means that the output of the pose estimation can be used in further processing and the whole system can be fine-tuned.

## 2.2 Human Activity Recognition (HAR)

Over the last decade, it has been distinguished that the process of HAR can be divided into four stages: capturing of signal activity, data pre-processing, AI-based activity recognition and the user interface for the management of HAR. Each stage has a multitude of techniques for its implementation, meaning the HAR system can be designed with multiple choices in mind. However, it is considered a challenging task due to the involvement of high level of abstraction and because the temporal dimension is not always easily handled. Previous approaches have deployed classical methods for features extractions [17, 18] where the central principal revolves around using the locations of body joints to select visual features in space and time. In recent years, 3D convolutions seem to be outputting the highest classification scores [19, 20] - the issue is that they require high number of parameters and need large amounts of memory for training. This can be improved by attention models which focus on body parts [21], with two-steam networks being utilised to merge both RGB images and the expensive optical flow maps [22]. Most 2D HAR methods use the body joint information only to extract localized visual features, as an attention mechanism. Hence, they may be limited only to datasets that provide skeletal data as methods exploring body joints cannot generate such features. Still, the scope for HAR is only increasing as deep learning (DL) methods such as deep neural networks (DNN), hybrid deep learning (HDL) models and transfer learning (TL) based models overtake ML ones due to superior performance. [23]

## 3 Methodology

In this section, I discuss the datasets used, along with the various methods and classifiers that have been used for predicting different human activity classes in them. Detailed descriptions have been provided for each approach used.

### 3.1 Datasets

One of the datasets used targets pose estimation which is derived from OpenPose. The motivation behind creating this dataset [24] is to be able to accurately predict human activities using an ML/DL model. It has been generated by running OpenPose on subjects performing different activities such as *'stand'*, *'walk'*, *'squat'* and *'wave'* in a video. By running OpenPose, the coordinates for crucial human key points such as *'nose_x'*, *'nose_y'*, *'LShoulder_x'* and *'LShoulder_y'* are provided which can essentially function as the input for my models. It works by initially pulling out features from the video using the first few layers and then inputting these extracted features into two parallel divisions of convolutional network layers. The first division predicts a set of 18 confidence maps - each denoting a specific part of the human pose skeleton – and the next branch predicts another set of 38 Part Affinity Fields (PAFs) which helps discern the degree of correlation between them. The later stages clean the predictions made, and with the help of the confidence maps, bipartite graphs are made between pairs of parts. Through PAF values, weaker links are pruned in the bipartite graphs. Applying all the given steps, estimations for human pose skeletons and data points are given from which the dataset has been made. It contains 37 columns, out of which 36 act as input and 1 ('class') as the target. All the input 36 are the (x,y) coordinates of the human key points in the referential videos.

The second dataset 'HARSENSE: STATISTICAL HUMAN ACTIVITY DATASET' [25] targets human activity recognition, specifically the daily living activity data, derived from inbuilt accelerometer and gyroscope sensors of the smartphones used (Poco X2 and Samsung Galaxy A32s). A total of 12 subjects have been used for the experiment, all above the age of 23 years and weighing more than 50 kilograms. The list of different activities (ADLs) are walking, standing, walking upstairs, walking downstairs, running and sitting. Data has been collected by mounting the smartphones on the waist and front pockets of the users, wherein all the activities were performed in a laboratory except for running, which has been performed on a football field. Hence, coordinates (x,y,z) relating to acceleration due to gravity, linear acceleration, gravity, rotational rate and rotational vector have been recorded and function as the input for my models, with the 'activity' column acting as the target. Although, data has been recorded individually for each user, the specific dataset that I use 'All Users Combined.csv", goes over the data points recorded for all the users. This is done in the hopes of creating models that would work on previously unseen users and users of varying types, rather than curating the models just for that user.

## 3.2 Model Approaches

In this sub-section I describe the classical models to solve the HAR and pose estimation problem, followed by the ensemble learning approaches.

### 3.2.1 Classical Models

**1. Logistic Regression**

Logistic regression is a predictive, supervised-learning method which models the relationship between a group of independent, multiclass variables and a binary variable. It predicts a binary outcome based on prior observations of a dataset and used a sigmoid function [26]. With this, it models the probability of human activities such as standing and walking based on the data. My model is optimized using the categorical cross-entropy loss function since the '*multi_class*' option is set to '*multinomial*' and the prediction involves a multi-class classification task of 6 classes for 1 dataset and 4 for the other. This is a task where any specific example can only belong to one out of various possible categories – the model decides which one it belong to. Hence, this loss function is appropriate in measuring the performance of a model since one example can be considered to belong to a category with probability 1 and other categories with probability 0. The 'solver' chosen is 'newton-fg' and there is no regularization as such.

**2. K-Nearest Neighbour (KNN)**

K-Nearest Neighbor (KNN) is a non-parametric, supervised-learning method that uses feature similarity, (i.e., the more closely out-of-sample features resemble the training set, the more likely they are to be classified to a certain group) to predict the values of any new or missing data. This means that new data points are assigned to different classes based on how closely they resemble other data points in the training set, with the algorithm making its predictions based on the majority of votes of its nearest neighbours. The algorithm utilizes a distance metric function such as Euclidean distance to find the k-nearest neighbours and classify the unknown data point into the most common class among its neighbours. Here, the '*k*' aspect of this algorithm represents the number of neighbours and is simply a hyperparameter that can be adjusted using a trial-and-error approach. For my model, I have set the number of neighbours as 30. [27]

### 3.2.2 Ensemble Learning Models

**1. Support Vector Machine (SVM)**

Support Vector Machines (SVM) are used for the classification of linear and nonlinear data. It utilizes a nonlinear mapping to convert the original training data into a higher dimension. It finds classification boundaries so that the classes are divided by a clear gap (margin) that is as wide as possible. Using this and support vectors, the algorithm outputs an optimal hyperplane which categorizes all new data points into different classes [27]. There are three commonly used kernel which are the linear kernel, radial basic function (rbf) kernel and the polynomial kernel. Although '*LinearSVC*' is more suited to larger datasets such as the selected 'HARSENSE' dataset, I have used '*SVC*' and set the '*kernel*' to '*rbf*' in my model. This is because there is a lack of linear correlation between data points in the dataset and hence my model requires a non-linear kernel, which is only catered in C-Support Vector Classification. In my model the regularization parameter (C) has been set as default but the kernel coefficient '*gamma*' has been set as '*auto*' in order to get the best performance.

**2. Random Forest**

Random Forest (RF) is a supervised-learning method that operates by constructing a large number of decision trees at training time (for my model the number of estimators or trees is set at the default value of 100). Since each individual tree in the random forest gives out a class prediction, the class that becomes the model's prediction is the mode of the classes of the individual trees [28]. RFs are not influenced by collinearity which is a reason why I have chosen this method. Here, another important parameter is the entropy used to train the model. I have set it as the default '*gini*' parameter for the measure of split quality due to the less computation required. This stands for the Gini index which is calculated by subtracting the sum of each class's squared probabilities from one, with larger partitions being favoured – they are also easier to implement [29].

## 3. Adaptive Boosting

An AdaBoost classifier is a meta-estimator that begins by fitting a classifier on the original dataset and then fits additional copies of the classifier on the same dataset but where the weights of incorrectly classified instances are adjusted such that subsequent classifiers focus more on difficult cases [30]. This works on the principle of boosting, which utilizes voting to integrate each individual model's output. It initializes by giving the same weight to all the instances in the training data, after which it calls the learning algorithm to create a classifier for this data and reweights each instance referring to the classifier's output. While the weight of all correctly classified instances is decreased, the weight of misclassified instances is increased. This generates a set of difficult instances with high weightage and easy instances with low weightage. In the succeeding iteration, a classifier is again constructed for the reweighted data which now, subsequently, focuses on classifying the difficult cases correctly. Then, according to the output of this new classifier, the weightage for each instance could again be changed [31]. With each iteration, the weights reverse how often the instances have been misclassified by the classifiers produced until now [32].

In my model, I have used a trial-and-error approach by manually testing out different number of estimators and learning rates. The results for the accuracies with all these combinations are shown below.

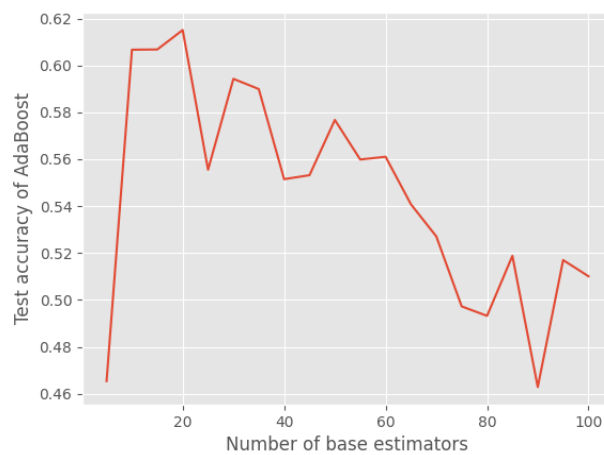

Figure 1: Illustration of accuracy v/s no. of base estimators in HAR dataset

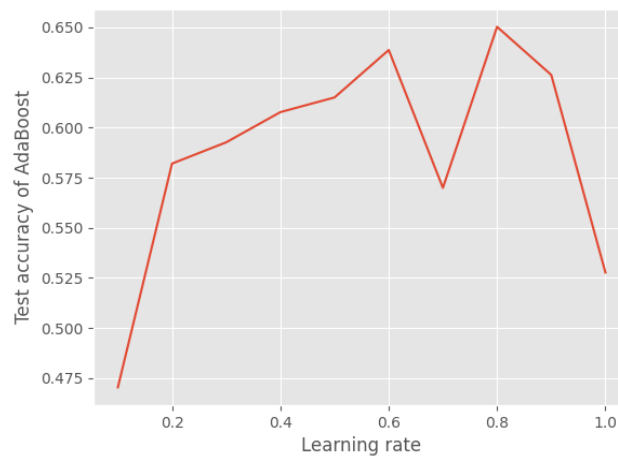

Figure 1: Illustration of accuracy v/s learning rates in HAR dataset

Seeing the results for myself, I have thus set '*n_estimators*' to 20 and '*learning_rate*' to 0.8, while using the default *DecisionTreeClassifier* as the '*base_estimator*' in order to achieve the best performance.

## 3.3 Implementation

Before training my models, pre-processing was required; so, both the datasets in consideration, were split by taking everything before the last index as training input and the string variable at the end as the target label (class). With this, the models were ready to be defined and so I did that by defining the training part and the testing part for each model as individual functions. The training function took parameters '*Xtrain*' (training input) and '*ytrain*' (training target labels), which would be used to fit each specific model and return the trained model; I also set the hyperparameters for every classifier with the aim of achieving the highest performance possible for each. Parallelly, the testing function took parameters '*clf*' (trained model), '*Xtest*' (testing input) and '*ytest*' (testing target labels), which would be used to return the list of the predicted values and the accuracy for the model. Here, the accuracy is calculated by first storing the difference between the predicted value and the actual value for each prediction. This is appended to a list after which all the non-zero values are counted and the total is divided by the length of the list; the result of this is subtracted from 1 and thus the accuracy is calculated.

With the declaration of these functions completed, I used the '*train_test_split()*' function from the data science library scikit-learn, to shuffle and split the data points for each class in the datasets into training subsets and test subsets that minimise the potential for bias in my evaluation and validation process. Now, with everything defined and prepared, it was actually time to run each model. This was done by calling all the functions defined for each model and using the output returned from '*train_test_split()*' function as their parameters. Through this, I printed the evaluation metrics such as accuracy, precision, recall and f-score for each model and can hence use this to compare the relative performance between them.

## 4 Results

In this section, I discuss the metrics used for evaluating the performance of each model and mention the results outputted with the help of tables and confusion matrixes.

### 4.1 Evaluation Metrics

**1. Accuracy**

Accuracy can be defined as the number of correct predictions divided by the total number of predictions, multiplied by 100. It is often the most common metric to judge a model but may not be a clear indicator of performance as it counts all of the true predicted values, but not specific for each label that exists. This is a serious concern if the goal is to predict a specific label, for example, a positive label, correctly, which in this case, it is. Higher accuracy doesn't mean that the model will have a good performance on predicting a specific label, hence the need for various, different performance metrics arises as included in this paper.

**2. Precision**

Precision is a metric that quantifies the number of correct positive predictions made. It, therefore, calculates the accuracy for the minority class. In binary classification, precision is the number of true positive results divided by the number of all positive results, including those not identified correctly. However, in an imbalanced classification problem with more than two classes, such as the one in this research paper, precision is calculated as the sum of true positives across all classes divided by the sum of true positives and false positives across all classes. The formula below is used to calculate the precision [33]:

Precision = Sum c in C TruePositives_c / Sum c in C (TruePositives_c + FalsePositives_c)

It can be seen as a measure of quality, with higher values indicating that an algorithm returns more relevant results than irrelevant ones.

**3. Recall**

Recall is a metric that quantifies the number of correct positive predictions made out of all positive predictions that could have been made. Unlike precision which only comments on the correct positive predictions out of all positive predictions, recall provides an indication of missed positive predictions. In this way, it provides some notice of the coverage of the positive class. In an imbalanced classification problem, such as the one in this research paper, it is calculated as the sum of true positives across all classes divided by the sum of true positives and false negatives across all classes. The formula below is used to calculate the recall [33]:

Recall = Sum c in C TruePositives_c / Sum c in C (TruePositives_c + FalseNegatives_c)

It can be seen as a measure of quantity with higher values meaning that an algorithm returns most of the relevant results (whether or not irrelevant ones are also returned).

**4. F-score**

The F-score or F-measure is a measure of a test's accuracy. It is calculated by taking mean of both precision and recall. Since, it takes both into account, a higher value for the f-score is better. It is calculated through the following formula:

$$\frac{2}{\frac{1}{precision} + \frac{1}{recall}} = \frac{2 \cdot precision \cdot recall}{precision + recall}$$

Looking at the formula, it can be seen that due to the product in the numerator, if one goes low, the final score also goes down significantly. Hence a model will do well, if the positive predicted are actually positives (precision) and doesn't miss out on positive and predicts them negative (recall).

**5. Confusion Matrix**

The confusion matrix, while not an exact evaluation metric, is a table that shows each class in the evaluation data and the number of correct predictions and incorrect predictions by comparing an observation's predicted class to its true class. It is used to visualize, in this case, the accuracy of multiclass classification predictive models like the ones used in this research paper [34].

**4.2 Pose Estimation Dataset**

Overall, the performance for all the models implemented with this dataset is excellent with high accuracies achieved by each ML method. Specifically, random forest stands out with a perfect accuracy of 100% and a perfect score for precision, recall and f-measure. Followed closely is AdaBoost's accuracy of 98.93% and KNN's accuracy of 98.23%. Both also have a perfect score for precision for 2 out of 4 classes, and at least 1 perfect score for recall too. The better performance in AdaBoost is showcased through perfect scores for f-measure; KNN, which although has scores above 95% for all classes, doesn't achieve any with 100%. Other models also perform relatively well, with none of the scores in any metric dropping below 90%. When considering the classes themselves, it can be seen that '*squat*' has a perfect score for precision while '*stand*' has a perfect score for recall.

**4.3 HAR Dataset**

Looking at the results for each metric, it is evident that only 3 out of the 5 models have performed excellently with the other 2 being subpar. Here, the highest accuracy of 96.71% is achieved by random forest, closely followed by SVM with an accuracy of 94.34% and KNN with an accuracy of 91.2%. The accuracy for AdaBoost is average at 65.03% while the accuracy for logistic regression doesn't even touch 50%, meaning that it will make only 1 in 2 correct predictions – an extremely poor result. Overall, the worst performing classes seem to be '*upstairs*' and '*downstairs*' with terrible precision, recall and f-measure scores achieved for each. Specifically, in logistic regression and AdaBoost, the f-measure score (which takes both precision and recall into account) is below 30% for the models, with the score for '*upstairs*' class in logistic regression even dipping to approximately 10%. It can also be seen, that many of the incorrect predictions made lie in '*walking*'. For example, in AdaBoost, out of 1056 data points for the '*upstairs*' class, 603 of the incorrect predictions made all lie in the '*walking*' class.

## 5 Discussion and Conclusions

In this research paper, I have approached the HAR and pose estimation problems using both classical and ensemble methods. While the classifiers perform relatively better for the OpenPose dataset than the HAR dataset, the common best classifier between them is random forest, a finding aligned with that of [35]. The reason behind this could be that since random forests uses bootstrap sampling and is not affected by multicollinearity, it does not make any assumptions about the underlying distribution of the data, allowing it to make more accurate predictions. On the other hand, we have logistic regression, which although performs extremely well for the OpenPose dataset, performs equally worse for the HAR one. The factor that could have caused this poor performance may be that there are little to no linear correlations between the class and input variables in the HAR dataset; since logistic regression only models linearly, it has thus performed poorly. I have made this assumption based on the results I was getting while implementing my models. Initially, I used the '*LinearSVC*' SVM method, which like logistic regression, models linearly. The accuracy I got with this was around 60%, but the moment I switched my model to '*SVC*' with a non-linear kernel, the accuracy jumped by more than 30%. Similarly, AdaBoost also performs extremely well on the OpenPose dataset, but terribly on the HAR one. This could be because the HAR dataset consists of a high noise level; however, to confirm this I would have needed to perform cross-validation and since I don't, it comes across as a limitation in this paper – a topic I discuss in the next paragraph. Given that AdaBoost, requires a quality dataset with little outliers [36], if the above is true, then the model would be prone to overfitting and hence perform terribly when tested with new, unknown data.

Equally important, there are a few shortcomings in the findings of this paper due to certain limitations. There are a number of elements that I have not explored – meaning there is quite a lot of potential for future research building upon this paper. The first way that could have helped me achieve better results is through hyperparameter tuning. By using approaches such as *GridSearchCV* or *RandomizedSearchVC* to parameter search, one can find the best arguments that should be passed into the classifier in reference. If I had adopted this, I may have achieved better accuracy and other evaluation metric scores for my implemented models. Another way is through using cross-validation, whose purpose is to test the ability of any ML model to predict new data. It can be used to flag problems like overfitting and selection bias, giving insights on how the model will generalize to a new subset of data. This is again, quite important, as discussed earlier, it may solve possible issues faced by the AdaBoost model. For it, cross-validation could be used to check any presence for noise and outline the bias that exists in the HAR dataset for the 'walking' class. Moreover, as it is quite evident that the models, especially logistic regression and Adaboost, favour this class while making predictions, solutions need to be found that can reduce the total number of instances of this class, while not ruining the general distribution of the dataset itself.

With this said, I believe that since my paper analyses the performance of different ML methods on activity recognition datasets, one being on pose estimation and the other on HAR, it can essentially act as a road map for success for industries or businesses pondering which classifiers to implement when designing their products. The range stretches from sports and fitness to surveillance and security, depicting the growing importance of activity recognition in real-world applications. This paper strives to express that, and although limited in its scope, can function as a base for further research and testing.

**Acknowledgements:**

∗Contact email: raviit.vij@gmail.com.
 I would like to thank Lumiere education for giving me this opportunity and allowing me to publish the research paper. I would also like to thank Prof. Owoeye Kehinde for his vital guidance and support while writing this paper. All remaining errors if any are my own.

# Appendix

**Appendix A: Accuracies**

Table 1: Accuracy summary table for OpenPose Dataset (%)

| ML Method | Accuracy |
|---|---|
| Logistic Regression | 97.53 |
| KNN | 98.23 |
| SVM | 96.82 |
| Random Forest | 100 |
| AdaBoost | 98.93 |

Table 2: Accuracy summary table for HAR Dataset (%)

| ML Method | Accuracy |
|---|---|
| Logistic Regression | 49.96 |
| KNN | 91.20 |
| SVM | 94.34 |
| Random Forest | 96.71 |
| AdaBoost | 65.03 |

**Appendix B: Precision**

Table 3: Precision summary table for each class in OpenPose Dataset (%)

| ML Method | 'stand' | 'walk' | 'squat' | 'wave' |
|---|---|---|---|---|
| Logistic Regression | 92.94 | 98.24 | 100 | 100 |
| KNN | 98.75 | 93.75 | 100 | 100 |
| SVM | 92.94 | 94.91 | 100 | 100 |
| Random Forest | 100 | 100 | 100 | 100 |
| AdaBoost | 100 | 98.33 | 100 | 97.40 |

Table 4: Precision summary table for each class in HAR Dataset (%)

| ML Method | 'walking' | 'standing' | 'upstairs' | 'downstairs' | 'running' | 'sitting' |
|---|---|---|---|---|---|---|
| Logistic Regression | 45.31 | 44.10 | 42.17 | 46.01 | 49.66 | 65.06 |
| KNN | 86.44 | 95.64 | 80.70 | 87.78 | 94.94 | 98.83 |
| SVM | 93.85 | 97.84 | 87.30 | 91.80 | 90.62 | 99.94 |
| Random Forest | 94.94 | 99.72 | 94.21 | 96.33 | 94.65 | 99.95 |
| AdaBoost | 53.82 | 86.30 | 28.31 | 37.80 | 80.33 | 83.54 |

**Appendix C: Recall**

Table 5: Recall summary table for each class in OpenPose Dataset (%)

| ML Method | 'stand' | 'walk' | 'squat' | 'wave' |
|---|---|---|---|---|
| Logistic Regression | 100 | 91.80 | 98.50 | 98.68 |
| KNN | 100 | 98.36 | 98.50 | 96.05 |
| SVM | 100 | 91.80 | 98.50 | 96.05 |
| Random Forest | 100 | 100 | 100 | 100 |
| AdaBoost | 100 | 96.72 | 100 | 98.68 |

Table 6: Recall summary table for each class in HAR Dataset (%)

| ML Method | 'walking' | 'standing' | 'upstairs' | 'downstairs' | 'running' | 'sitting' |
|---|---|---|---|---|---|---|
| Logistic Regression | 70.62 | 63.45 | 5.87 | 15.25 | 19.44 | 74.46 |
| KNN | 95.49 | 99.32 | 78.03 | 71.55 | 84.68 | 99.95 |
| SVM | 96.25 | 99.52 | 82.67 | 77.91 | 96.15 | 99.84 |
| Random Forest | 98.24 | 100 | 89.49 | 88.24 | 96.35 | 100 |
| AdaBoost | 72.37 | 73.78 | 26.70 | 14.88 | 53.94 | 99.43 |

**Appendix D: F-Score**

Table 7: F-score summary table for each class in OpenPose Dataset (%)

| ML Method | 'stand' | 'walk' | 'squat' | 'wave' |
|---|---|---|---|---|
| Logistic Regression | 96.34 | 94.92 | 99.25 | 99.38 |
| KNN | 99.37 | 96.00 | 99.25 | 97.99 |
| SVM | 96.34 | 93.33 | 99.25 | 97.99 |
| Random Forest | 100 | 100 | 100 | 100 |
| AdaBoost | 100 | 97.52 | 100 | 98.04 |

Table 8: F-score summary table for each class in HAR Dataset (%)

| ML Method | 'walking' | 'standing' | 'upstairs' | 'downstairs' | 'running' | 'sitting' |
|---|---|---|---|---|---|---|
| Logistic Regression | 55.20 | 52.03 | 10.30 | 22.90 | 27.94 | 69.45 |
| KNN | 90.74 | 97.45 | 79.34 | 78.84 | 89.51 | 99.39 |
| SVM | 95.03 | 98.68 | 84.92 | 84.28 | 93.30 | 99.90 |
| Random Forest | 96.56 | 99.86 | 91.79 | 92.10 | 95.49 | 99.97 |
| AdaBoost | 61.73 | 79.56 | 27.48 | 21.36 | 64.55 | 90.80 |

**Appendix E: Confusion Matrix**

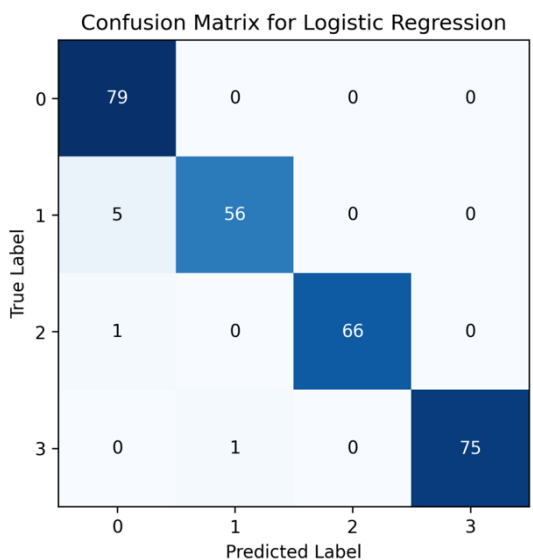

Figure 3: Multi-class confusion matrix for Logistic Regression model in OpenPose Dataset

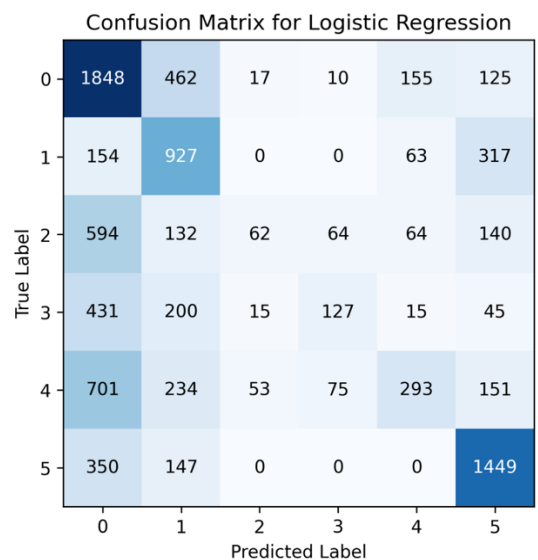

Figure 4: Multi-class confusion matrix for Logistic Regression model in HAR Dataset

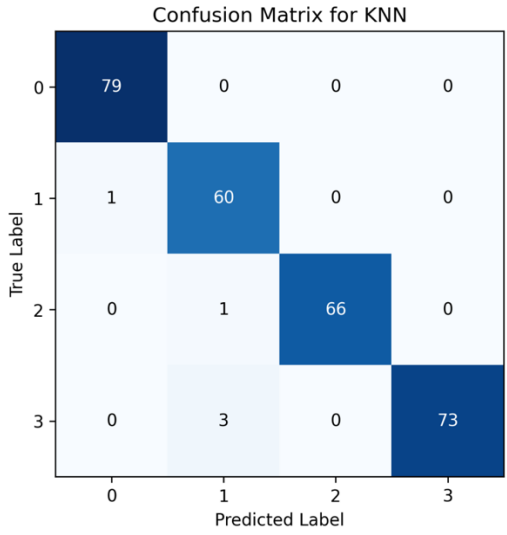

Figure 5: Multi-class confusion matrix for KNN model in OpenPose Dataset

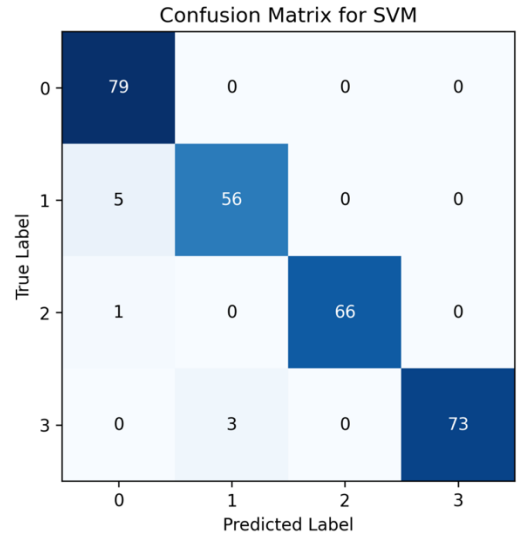

Figure 7: Multi-class confusion matrix for SVM model in OpenPose Dataset

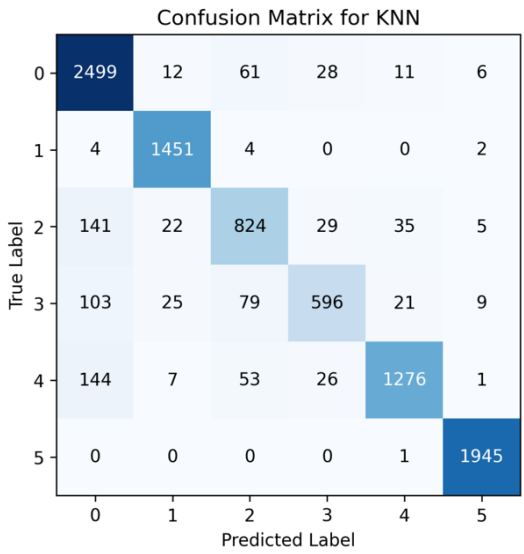

Figure 6: Multi-class confusion matrix for KNN model in HAR Dataset

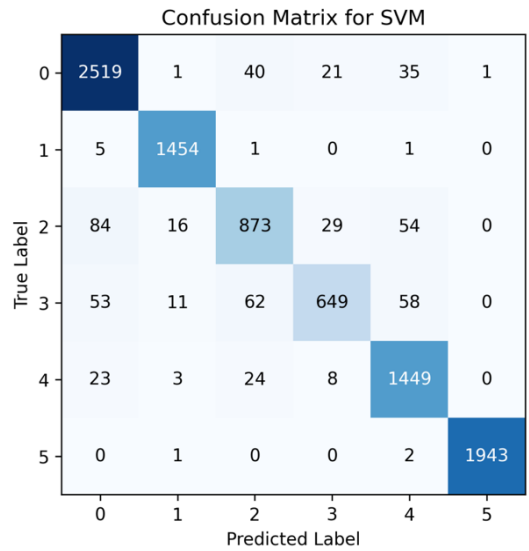

Figure 8: Multi-class confusion matrix for SVM model in HAR Dataset

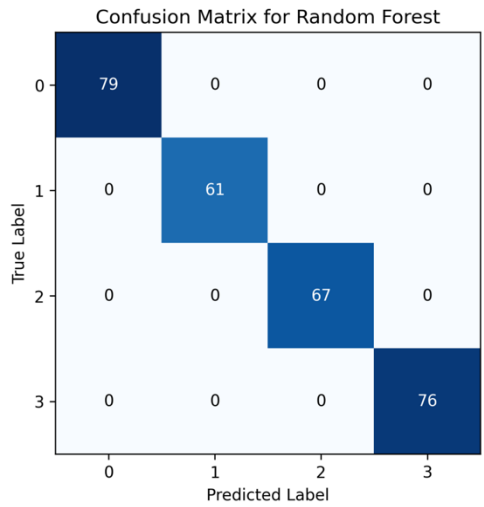

Figure 9: Multi-class confusion matrix for Random Forest model in OpenPose Dataset

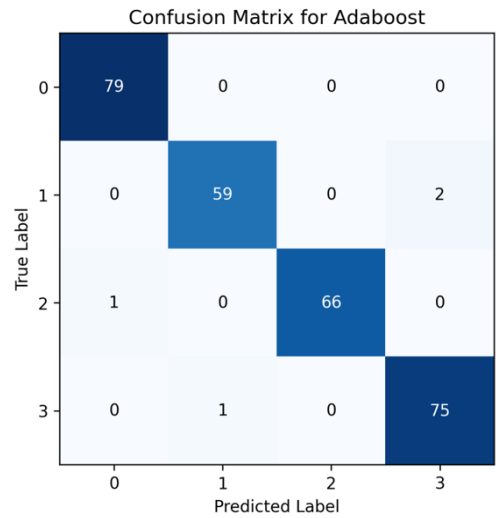

Figure 11: Multi-class confusion matrix for AdaBoost model in OpenPose Dataset

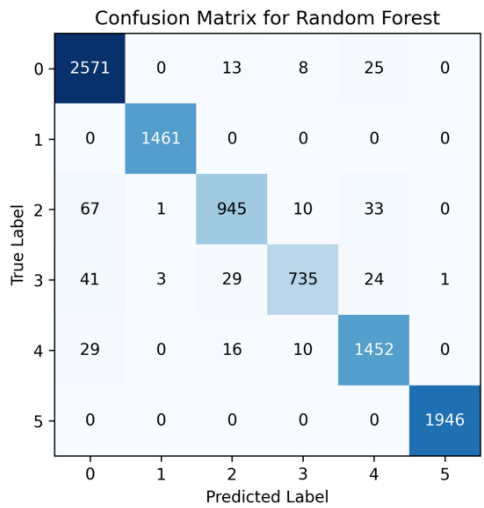

Figure 10: Multi-class confusion matrix for Random Forest model in HAR Dataset

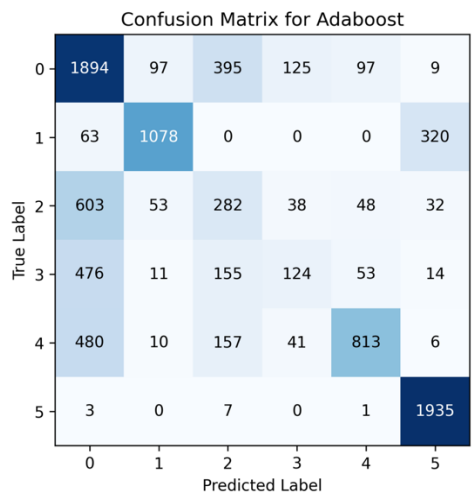

Figure 12: Multi-class confusion matrix for AdaBoost model in HAR Dataset